%% file: main.tex
\documentclass[10pt]{llncs}
\usepackage{rotating}
\usepackage{pdflscape}
\usepackage{amsmath}
\usepackage{amsfonts}
\usepackage{amssymb}
\usepackage{amstext}
\usepackage{latexsym,stmaryrd}
\usepackage[ruled, linesnumbered,noend]{algorithm2e}

\usepackage{inferance}
\usepackage{deduc}

\usepackage{xcolor}

\input{symbols.tex}

\renewcommand\ruleformat[1]{\ #1}

\begin{document}
\title{ Extracting Frequent Gradual Patterns Using Constraints Modeling}
%\title{A SAT-Based Approach for Extracting Frequent Gradual Patterns}
%\title{Mining Frequent Gradual Patterns Using Boolean Satisfiability}
\titlerunning{Mining Gradual Patterns Using SAT}
\author{Jerry Lonlac$^{1,2}$, Said Jabbour$^{2}$, Engelbert Mephu Nguifo$^{3}$, Lakhdar Sais$^{2}$, Badran Raddaoui
$^{4}$}

\institute{
$^1$ Centre de Recherche, IMT Lille Douai, Universit\'e Lille, Lens, France \\
\{jerry.lonlac\}@imt-lille-douai.fr \\
$^2$ Univ. Lille-Nord de France, CRIL/CNRS UMR 8188 Lens France \\
\{jabbour, sais\}@cril.fr \\
$^3$ Univ. Clermont Auvergne, CNRS, LIMOS, F-63000 Clermont-Ferrand, France \\
%\{mephu\}@isima.fr
 engelbert.mephu\_nguifo@uca.fr \\
%$^2$ University Clermont Auvergne, CNRS, GEOLAB, F-63000 Clermont-Ferrand \\
$^{4}$ SAMOVAR, T\'el\'ecom SudParis, CNRS, Univ. Paris-Saclay, F-91011 Evry Cedex, France \\
badran.raddaoui@telecom-sudparis.eu
}

\maketitle

 \begin{abstract}
In this paper, we propose a constraint-based modeling approach for the
problem of discovering frequent gradual patterns in a numerical dataset. 
This SAT-based declarative approach offers an additional possibility to benefit from the recent progress in satisfiability testing and to exploit the efficiency of modern SAT solvers for enumerating all frequent gradual patterns in a numerical dataset. 
Our approach can easily be extended with extra constraints, such as temporal constraints in order to extract more specific patterns in a broad range of gradual patterns mining applications.
%The experimental evaluation on real world datasets shows the feasibility of our proposed approach in practice.
We show the practical feasibility of our SAT model by running experiments on two real world datasets.

\end{abstract}
 
\section{Introduction}

Frequent pattern mining is a well-known and essential problem of data mining. Its goal is to efficiently discover in large volumes of data the hidden patterns having more occurrences than a pre-defined threshold. This problem started with the analysis of transactional data \cite{AgrawalS94} (frequent itemsets), and quickly expanded to the analysis of data having more complex structures such as sequences, trees or graphs. A new pattern mining problem has then been introduced in transactional databases where attributes can have a numeric value:  mining frequent gradual itemsets (also known as gradual patterns). Frequent gradual patterns mining problem consists to discover frequent co-variations between numerical attributes in Databases \cite{AyouniLYP10}, such as: “The higher the age, the higher the salary”. This problem has been tackled since many years \cite{SrikantA96} and find the numerous applications for the numerical databases such that the biological and medical databases. Several algorithms have been
proposed in the literature in order to address this problem. Most of theses algorithms use very often the data mining algorithms for automatically extract  the gradual patterns. \cite{Hullermeier02,BerzalCSVS07,Masseglia08gradualtrends,Di-JorioLT08,Di-JorioLT09,LaurentLR09}. 

In \cite{Hullermeier02}, for extracting such patterns, the authors apply a linear regression analysis between pairs of attributes and the validity of the gradual tendency between two attributes is evaluate from the quality of regression.  This validity is measured by the normalized mean squared error $R^2$, together with the slope of the regression line: attribute pairs that are insufficiently correlated are rejected, as well as pairs for which one attribute remains almost constant while the other one increases, which can be detected by a low slope of the regression line. In \cite{BerzalCSVS07}, the authors use for the first time the data mining approaches from a adaptation of \texttt{Apriori} algorithm for extracting gradual patterns. They formulate the  problem of extraction of such patterns as the problem of discovery classical itemset in a suitable set of transactions $\Delta'$ obtained from the initial data set $\Delta$. Each pair of objects in the initial data is associated to a transaction in the derived data set $\Delta'$; each item $I$ of $\Delta$ defines two items  $I^{\leq}$ and $I^{\geq}$ in $\Delta'$ instead one item. A transaction $t$ in $\Delta'$ then possesses an item $I^*$ $(* \in \{\geq, \leq \})$ if the pair of objects $(x, x')$ of $\Delta$ satisfies the constraint imposed by $I^*$, i.e. $A(x) * A(x)$. Formulate thus, a gradual pattern in $\Delta$ is equivalent to a classical itemset extracted from $\Delta'$. The computing of support of gradual pattern makes this approach complex. In fact, the support is compute by considering all object couples and explicitly building the data set $\Delta'$ to apply a classic frequent itemset mining algorithm have too high a computational cost.

Whereas the proposed algorithm in \cite{BerzalCSVS07} are limited to six attributes, in \cite{Di-JorioLT09} a first efficient algorithm for mining gradual itemsets and gradual rules capable of handling databases with hundreds of attributes was proposed. One of the major problem of the mining gradual pattern approaches is the exponential combination space to explore and the problem of handling the quantity of extracted patterns which can be also of exponential size. This combinatorial explosion is tackled in \cite{LaurentLR09,AyouniLYP10,DoTLNTA15}. In fact, in \cite{LaurentLR09}, the authors propose an approach for extracting gradual patterns from large datasets which takes advantage of a binary representation of lattice structure. In \cite{AyouniLYP10,DoTLNTA15}, in order to reduce the quantity of patterns, it is proposed to mine only the closed frequent gradual patterns.

Our approach for extracting all frequent closed gradual patterns in a numerical dataset differs from all the previous specialized approaches. It follows the SAT-Based framework  proposed  in \cite{JabbourSS13} for mining frequent closed itemsets.
This new framework offers a declarative and flexible representation model. New constraints often require new implementations in specialized approaches, while they can be easily integrated in such Boolean satisfiability framework. It allows data mining problems to benefit from several generic and efficient SAT solving techniques.
In \cite{CoqueryJSS12}, the authors show how some typical constraints (e.g. frequency, closure, monotonicity) used in itemset mining can be formulated for use in Boolean satisfiability. This first study leads to the first SAT approach for itemset mining displaying nice declarative opportunities without neglecting efficiency.
Considering the promising results obtained from this framework, we propose to heavily exploit the declarative language  Boolean satisfiability (SAT) and these associated efficient and generic solving techniques. First, we propose a new SAT model for the problem of mining frequent gradual pattern that includes different types of constraints.
%by taking account the cases where several transactions have the same value for an attribute. 
We propose the new constraints different from those proposed in \cite{CoqueryJSS12}. 
The first one, encodes that a gradual itemset not must contain both a gradual item and its complementary gradual item.
The second one, allows us for a given gradual itemset $s$,  to place uniquely one transaction in each position of the longest sequence of transactions respecting $s$.  
The third constraint captures the fact that a transaction should not be placed in more than one position of the longest sequence of transactions respecting $s$.
The others constraints allow to link the gradual item with the transactions in order to compare the transactions with respect to gradual item of a given gradual itemset. This link allows us to detect for a given gradual itemset $s$,  all the sequences of transactions which respect $s$. 
Any gradual itemset, representing co-variations items has a symmetric gradual itemset where the items are the same and the variations are all reversed. If a gradual itemset is frequent, then its symmetric gradual itemset is also frequent and does not carry additional information. 
This symmetry of problem allows to generate only half of gradual itemsets for automatically deduce the other ones. We take into account this symmetry by forcing the SAT solver during the search process to affect always a positive polarity to the first variable selected by branching heuristic at each restart.

%The gradual itemset are symmetric, in fact, if a gradual itemset is frequent, then a symmetric gradual itemset is also frequent and does not carry additional information. This symmetry of the problem allows to generate only half of the gradual itemsets for automatically deduce the other ones. We take  into account this symmetry during the resolution  process of our obtained boolean formulation of constraints by branching at each restart of our solver on a positif literal. 

%This constraint encodes that the equal values of transactions of the same attribute meaning that we have neither increasing and neither decreasing of values of this attribute between the transactions considered. It must note that,  the problem of equal values evoked for the first time in \cite{DoTLNTA15}, appears very often in real datasets.

Finally, we provide a boolean formulation of closedness  constraints, in order to search for frequent closed gradual patterns. This allows us to obtain a SAT-based model for enumerating frequent closed gradual patterns in a numerical dataset.
% including an interesting encoding of the equal values constraint of a attribute to a Boolean formula.

The paper is organized as follows: in Section \ref{sec:gradual}, we present the problem of mining gradual itemsets from numerical databases and some efficient algorithms proposed in the data mining domain for the automatic extracting such patterns. We also recall the definition of Boolean satisfiability problem, called SAT. Section \ref{enumSAT} describes the SAT-Based enumeration procedure to deal with the problem of enumerating all models of a CNF formula.
In Section \ref{encodingSAT}, we describe our SAT encoding of frequent gradual itemset mining problem and show through an example how it can be applied to find frequent gradual itemsets in a numerical dataset. Finally, section \ref{exps} presents detailed experiments carried out over real datasets, showing the applicability  and the interest of our approach.
%We conclude and give some perspectives in Section 5.

%%%%%%%%%%%%%%%%%%%%%%%%%%%%%%%%%%%%%%%%%%%%%%%%%%%%%%%%%%%%%%%%%%%%%
\section{Preliminaries}
\label{sec:gradual}
In this section, we formally describe the problem of mining frequent gradual itemsets (patterns) and the mining closed gradual patterns in a numerical dataset. We then present some approaches of state of the art proposed to automatically extract such patterns. We also recall the boolean satisfiability problem, commonly called SAT and the definitions of gradual patterns given in \cite{Di-JorioLT09}.

\subsection{Gradual patterns mining problem}
%\label{sec:gradual}

The problem of mining gradual patterns consists in mining attribute co-variations in numerical dataset of the form \textit{"The more/less X, . . . , the more/less Y"}. %(Ayouni et al. 2010)
We assume here that we are given a dataset $\Delta$ containing a set of objects $\cal{T}$ that defined a relation on a attribute set $\cal{I}$ with numerical values $\cal{I}$ i.e. for
$t  \in \cal{T}$, $t[i]$ denotes the value of the attribute $i$ over object $t$.

For instance, we consider the numerical dataset given in Table \ref{tab:rel1} which gives for each date, the quantity of each species presents in a given ecosystem. This table contains eight objects ($\{t_1, .. ., t_8 \}$) and three  attributes represents by the  scientific names of the different species (\textit{Poaceae, Secale, Rumex}).

 \begin{table}
\center\small
\begin{tabular*}{\linewidth}{@{\extracolsep{\fill}}lccc}
  \hline
  \textbf{Dates} & \textbf{Poaceae (p)} & \textbf{Secale (s)} & \textbf{Rumex (r)} \\
\hline 
 t1 & 4 & 3 & 13 \\
\hline
 t2 & 6 & 9 & 11  \\
  \hline
 t3 & 8 & 1 & 9  \\
\hline
t4 & 13 & 7 & 5 \\
\hline
t5 & 4 & 5 & 10 \\
\hline
t6 & 9 & 6 & 8  \\
\hline
t7 & 10 & 6 & 12  \\
\hline
t8 & 13 & 7 & 13  \\
\hline
\end{tabular*}
\vspace{0.2cm}
  \caption{Abundance of species in a Ecosystem}
  \label{tab:rel1}
\end{table}

%Three kinds of variations have to be considered: increasing variation, decreasing variation, and no variation. 
Each attribute will hereafter be considered twice: once to indicate its increasing, and once to indicate its decreasing, using the $\leq$ and $\geq$ operators. This leads to consider new kinds of items, called gradual items. 

\begin{definition}[Gradual Item]
\label{def:gradualItem}
Let $\Delta$ be a dataset defined on a numerical attribute set $\cal{I}$, A gradual item is defined under the form $i^*$, where $i$ is a attribute of $\cal{I}$ and $* \in \{\geq, \leq\}$ be a comparison operator.
\end{definition}

If we consider the numerical dataset given in table \ref{tab:rel1}, $Poaceae^{\geq}$ (respectively $Poaceae^{\leq}$) is a gradual item meaning that the values of attribute \textit{Poaceae} is increasing (respectively decreasing).

A gradual itemset (gradual pattern) is then defined as follow:

\begin{definition}[Gradual Itemset]
\label{def:gradualItemset}
A gradual itemset $s = (i_1^{*_1}, ... , i_k^{*_k})$ is a non
empty set of gradual items.  A gradual $k$-itemset is an gradual itemset containing $k$ gradual items. 
\end{definition}

For example, $\{Poaceae^{\geq},  Rumex^{\leq}\}$ is a gradual itemset meaning that \textit{"more the values of attribute Poaceae increase, more the values of attribute Rumex decrease"}.

A gradual itemset imposes a variation constraint on several
attributes simultaneously. The length of a gradual itemset is equal to  number of gradual item that it contains. ($Poaceae^{\geq},  Rumex^{\leq}$) is a gradual $2$-itemsets.

The support (frequency) of a gradual itemset in a dataset amounts to the extent to which a gradual pattern is present in a given database. Several support definitions have been proposed in the literature, showing that gradual itemsets can follow different semantics. The choice between them generally depends on the considered application. 

A gradual itemset is said to be frequent if its frequency is greater than or equal to a user-defined threshold. 

%The problem of mining frequent gradual itemsets is to find the complete set of frequent gradual itemsets in a given dataset $\Delta$ containing numerical attributes, with respect to a minimum threshold \textit{minSupp}.

\begin{definition}[Frequent Gradual Itemsets Mining Problem]
Let $\Delta$ be a numerical dataset and \textit{minSupp} a minimum  support threshold.
The problem of mining gradual itemsets is to find the set of all frequent gradual itemsets of $\Delta$ with respect to \textit{minSupp}.
%i.e. finding the set $\{g| Supp(g, \Delta)\geq\lambda\}$.
\end{definition}

In the following section, we present the different semantics and algorithms that have been proposed to automatically extract gradual itemsets from numerical dataset.

\subsection{Discovering frequent gradual patterns}
\label{sec:Mining_gradual}
Gradual patterns can be compared to fuzzy gradual rules that have first been used for command systems some years ago \cite{DuboisP92a,DuboisPU03}, for instance for braking systems: “the closer the wall, the stronger the brake force”. Whereas such fuzzy gradual rules are expressed in the same way as the gradual patterns, the main difference is that fuzzy gradual rules were not discovered automatically from data. They were designed by human experts and provided as input to expert systems.

Several works in the pattern mining field have shown that it was feasible to mine automatically such rules from raw data  \cite{BerzalCSVS07,Hullermeier02,Di-JorioLT09}. However, the quantity of mined patterns (and, consequently, the quantity of extracted rules) makes their exploitation difficult. So, as mentioned above, in \cite{AyouniLYP10,DoTLNTA15}, the authors propose to mine only closed gradual patterns in order to reduce the to reduce the number
of patterns extracted without loss of information. 
This preliminary work didn't exploit closure properties to improve the mining algorithm and reduce execution time. 
However, mining gradual patterns is a costly task in terms of computation time. It was proposed in \cite{LaurentNST10a} to exploit the parallel processing capabilities of multi-core architectures in order to reduce computation time. 
%The main contribution of the present work is to provide a boolean satisfiability encoding of the problem of mining gradual patterns by considering the gradual patterns definition used in the association rule formulation \cite{BerzalCSVS07} which take account the cases where several transactions have the same value for an attribute. We then exploit the scalability of modern SAT solvers to discover frequent closed gradual patterns from the models of the obtained boolean formula. 

The evaluation of the support of gradual patterns has been defined in different manners depending on the semantic  and the application considered. In \cite{Hullermeier02}, it is based on regression, while \cite{BerzalCSVS07} and \cite{LaurentLR09} consider the number of transactions that are concordant and discordant, in the idea of exploiting the Kendall’s tau ranking correlation coefficient \cite{KendallB39}. This means that given a gradual itemset $s$, all pairs of transactions $(t_i, t_j )$ will be compared according to the order induced by $s$, and the support will be based on the proportion of these pairs that satisfy all gradual items in $s$. The interest of this definition is that it makes possible to take into account the amplitude of the distortion for data that do not satisfy the gradual patterns.

In contrast, the definition of support proposed in \cite{Di-JorioLT09} is based on the length of the longest sequence of transactions that can be ordered consecutively according to a gradual pattern $s$. The interest of this definition is that such transaction sequences can then be easily presented to the analyst, allowing to isolate and reorder a part of data and to label it with a description in terms of co-variations (the gradual itemset being this description).

The main contribution of the present work is to provide a boolean satisfiability encoding of the problem of mining frequent gradual patterns by considering the gradual patterns definition used in the association rule formulation \cite{Di-JorioLT09}. We then exploit the scalability of modern SAT solvers to discover frequent closed gradual patterns from the models of the obtained boolean formula.

In this paper, For a given attribute $i$ in a dataset $\Delta$, we consider two gradual items $i^{\leq}$ and $i^{\geq}$, as consider in the algorithms of extracting frequent gradual itemsets  proposed in the data mining domain \cite{Di-JorioLT08,Di-JorioLT09,LaurentLR09}. 
We use the variation semantic proposed by \cite{Di-JorioLT09}, which defined the support of a gradual itemset as being the maximum number of transactions (the size of the longest sequences of transactions) that can be ordered w.r.t. this gradual itemset.
In order to explain this semantic, we present first the definition of the order induced by a gradual itemset \cite{DoTLNTA15}.

\begin{definition} [Gradual itemset induced order]
Let $s = (i_1^{*_1}, ... , i_k^{*_k})$ be a gradual itemset , and $\Delta$ be a numerical dataset. Two objects $t$ and $t'$ of $\Delta$ can be ordered w.r.t. $s$ if all the values of the corresponding items from the gradual itemset can be ordered to respect all the variations of the gradual items of $s$ : for every $l \in [1, k]$, $t[i_l] \leq t'[i_l]$ if $*_l = \geq$ and $t[i_l] \geq t'[i_l]$
if $*_l = \leq$. The fact that $t$ precedes $t'$ in the order induced by $s$ is denoted $t \lhd_s t'$.
\end{definition}

Referring to the previous example from Table \ref{tab:rel1}, $t_1$ and $t_2$ can be ordered with respect to gradual itemset $s_1 = (Poaceae^{\geq},  Rumex^{\leq})$ as $t_1[Poaceae] \leq t_2[Poaceae]$ and $t_1[Rumex] \geq t_2[Rumex]$: we have $t_1 \lhd_{s_1} t_2$.

This order is only a partial order. For example consider $t_2$ and $t_5$ of Table \ref{tab:rel1} : they can't be ordered according to $s_1$.
In fact, the pattern $s_1$ is not relevant to explain the variations between $t_2$ and $t_5$ , and more generally all transaction pairs that it can't order. Conversely, a gradual pattern is relevant to explain the  variations occurring in the transactions that it can order. The support definition that we consider in this paper for our encoding SAT goes further and focuses on the size of the longest sequences of objects that can be ordered according to a gradual itemset. The intuition being that such patterns will be supported by long continuous variations in the data (long periods of co-evolution between paleoecological indicators in the case of paleoecological data given by table \ref{tab:rel1}), such continuous variations being particularly desirable to extract in order to better understand the data.

\begin{definition} [Support of a Gradual Itemset]
Let $\Delta$  be a numerical dataset containing a set of objects $\{t_1, ... , t_n \}$ and $L = \langle t_{i_1}, ... , t_{i_s} \rangle$ be a sequence of objects from $\Delta$, with $\forall k \in [1..s], i_k \in [1..n]$ and $\forall k, k' \in [1..s], k \neq k' \Rightarrow i_k \neq i_k'$. Let $s$ be a gradual itemset. $L$ respects $s$ if $\forall k \in [1..s-1]$ we have $t_{i_k} \lhd_s t_{i_{k+1}}$. Let $L_s$ be the set of objects that respect $s$. The support of $s$ is define by $Support(s) = \frac{max_{L \in L_s}(|L|)}{|\Delta|}$. i.e. it is the size of the longest list of tuples that respects $s$.
\end{definition}

Note that the support of a gradual itemset containing a single gradual item is always $100\%$ as it is always possible to order all the tuples by one column.

%\textcolor{red}{
By considering the dataset of table \ref{tab:rel1} and the pattern $s_1 =$ $(Poaceae^{\geq},  Rumex^{\leq})$, the set of all the lists of sequence of objects respecting $s_1$ is $L_{s_1}$ $= \{  \langle t_1, t_2, t_3, t_6, t_4\rangle$,  $\langle t_1, t_5, t_3, t_6, t_4 \rangle,$  $\langle t_1, t_7, t_4\rangle,$  $\langle t_1, t_8, t_4\rangle \} $.
Two lists from $L_{s_1}$ have a maximal size, which is 5. Hence, $support(s_1) = \frac{5}{8} = 0.625$, meaning that $62.5\%$ of the input objects can be ordered consecutively according to $s_1$. 
%}

%The gradual item $i^o$ represents the cases where the objects of dataset have the same value for attribute $i$. This will allow us to also solve the no trivial problem of equal values \cite{DoTLNTA15}.
%We consider the variation semantic proposed in \cite{BerzalCSVS07}, which defined the support of a gradual itemset as being the proportion of objects couples of dataset that verify the constraints expressed by all the gradual items in the itemset.

\begin{definition}[Complementary Gradual Itemset]
\label{def:patternCompl}
Let $s = (i_1^{*1}, ..., i_k^{*k})$ be a gradual itemset, and $c$ be a function such that $c(\geq) = \leq$ and $c(\leq) = \geq$. Then $c(s) = (i_1^{*^c_1}, ..., i_k^{*^c_k}) $  is the complementary (symmetric) gradual itemset of $s$ and is defined as $\forall j \in [1..k], *^c_j = c(*_j)$.
\end{definition}

The complementary  gradual itemset (symmetric gradual itemset) of $s_1$ is denoted  $c(Poaceae^{\geq},  Rumex^{\leq}) = (Poaceae^{\leq},  Rumex^{\geq})$.

\begin{proposition}
\label{pro:compl}
$Support(s) = Support(c(s))$.
\end{proposition}

The proposition \ref{pro:compl} given in \cite{Di-JorioLT09} avoids unnecessary computations, as generating only half of the gradual itemsets is sufficient to automatically deduce the other ones. This means that for each gradual itemset there is a symmetric gradual itemset having the same support.

\subsection{Closed gradual patterns}

In data mining, closed patterns are key to obtain a condensed representation of the patterns without loss of information \cite{PasquierBTL99}.
An pattern $I$ is said closed if there is no pattern $I'$ such that $I \subset I' $ and $support(I') = support(I)$.
This notion of closure has been  introduced for the first time in the gradual patterns in \cite{AyouniLYP10} where the author propose an pair of functions $(f,g)$ defining a Galois closure operator for gradual patterns.

Given a set of sequence of transactions $\cal{L}$ of a dataset, the function $f$ returns the gradual pattern $s$ (all the attributes (items) associated with their respective variations) respecting all transaction sequences in $\cal{L}$. While the function $g$ returns for a given gradual pattern $s$ the set of the maximal sequences of transactions $\cal{L}$ which respects the variations of all gradual attribute in $s$.

Provided these definitions, a gradual pattern $s$ is said to be closed if $f(g(s)) = s$. In \cite{AyouniLYP10}, the authors use these definitions rather as a post-processing step. In \cite{DoTLNTA15}, these definitions are included in the mining process and allow to benefit from the runtime and memory reduction.

Let us consider the dataset of the Table \ref{tab:rel1}. Thus, we have for example : \\ $g(\{Poaceae^{\geq},Rumex^{\leq}\}) = \{\langle t_1, t_2, t_3, t_6, t_4 \rangle, \langle t_1, t_5, t_3, t_6, t_4 \rangle \}$ and  \\ $f(\{\langle t_1, t_2, t_3, t_6, t_4 \rangle, \langle t_1, t_5, t_3, t_6, t_4 \rangle \}) = \{Poaceae^{\geq} ~ Rumex^{\leq}\}$. Therefore, \\ $\{Poaceae^{\geq},Rumex^{\leq}\}$ is a closed gradual pattern.

Compared to the context of classical items, the main issue here is to manage the fact that the function $g$ does not return a set of transactions but it returns a set of sequences of transactions. We propose below a new SAT-based approach for discovering frequent closed gradual patterns in the numerical dataset. 
%This approach  extends the approach proposed in \cite{BerzalCSVS07} and propose a simple solution to take into account equal values during the mining process. In fact, the approach proposed in \cite{BerzalCSVS07} no take into account equal values and is limited to the extraction of gradual patterns of length 4.
Our proposed approach allows to extract all the frequent gradual patterns with respect to the minimum support threshold by benefiting from the impressive progress in boolean satisfiability checking \cite{BiereMH2009} and from the scalability of modern SAT solvers.
%=====================================================================
\subsection{Boolean satisfiability}
In this section, we introduce the Boolean satisfiability problem, called SAT. It corresponds to the problem of deciding if a formula of propositional classical logic is consistent or not. It is one of the most studied NP-complete decision problem. In this work, we consider the associated problem of boolean model enumeration.

We consider the conjunctive normal form \textit{(CNF)} representation for the propositional formulas. A \textit{(CNF)} formula $\cal{F}$ is a conjunction of clauses, where a \textit{clause} is a disjunction of literals. A \textit{literal} is a positive $(l)$ or negated $(\neg l)$
propositional variable. The two literals $(l)$ and $(\neg l)$ are called
complementary.
We note by $\bar{l}$  the complementary literal of $l$.
For a set of literals $L$, $\bar{L}$ is defined as $\{\bar{l} ~|~ l \in L\}$.

%A \textit{(CNF)} formula can also be seen as a set of clauses, and a
%clause as a set of literals. 
Let us recall that any propositional formula can be translated to \textit{(CNF)} using linear Tseitin's encoding \cite{Tseitin68}. 
The set of variables occurring in ${\cal F}$ is noted $Var(\cal{F})$.

An {\it interpretation} $\rho$ of a boolean formula ${\cal F}$ is a function which associates a value $\rho$($l$)$\in\{0, 1\}$ (0 correspond to false and 1 to true) to the variables $x \in Var(\cal{F})$.
A model of a formula is an {\it interpretation} $\rho$  that satisfies the formula. SAT problem consists in deciding if a given formula admits a model or not.
%We note by $\bar{l}$  the complementary literal of $l$.
%For a set of literals $L$, $\bar{L}$ is defined as $\{\bar{l} ~|~ l \in L\}$.

\section{SAT-Based enumeration procedure}
\label{enumSAT}
In this, we describe the SAT-Based enumeration procedure to deal with the problem of enumerating all models of a CNF formula. SAT is a decision problem. When the answer is positive, the current SAT solvers provide a model satisfying the formula.
In the sequel, we briefly describe the basic components of moderns 
SAT solvers so call CDCL SAT solvers \cite{Moskewicz01,EenS03} designed to enumerate all the models of a given CNF formula.
To be exhaustive, these solvers incorporate unit propagation (enhanced by efficient and lazy data structures), variable activity-based heuristic, literal polarity phase, clause learning, restarts and a learnt clauses
database reduction policy.

\vspace{0.3cm}
%\restylealgo{ruled}\linesnumbered

%\restylealgo{ruled}\linesnumbered

\begin{algorithm}[h]
  \caption{CDCL Based Enumeration solver}
  \label{algoExtendCDCL}
  %\SetVline
  \SetInd{0.3em}{0.7em}
  \KwIn{ a CNF formula $\cal{F}$}
  \KwOut{ All models of $\cal{F}$}
    
  \SetKwBlock{Begin}{Debut}{Fin}  
  {
   % \If{\TIMEREDUCE{}}{
      $\rho = \emptyset$  \tcc*{interpretation}
      $\delta = \emptyset$   \tcc*{learnt clauses database}
      $dl = 0$   \tcc*{decision level}

      \While{$(true)$}
      {
      	Prop \;
        $\gamma = unitPropagation(\cal{F,I})$ \;
        \If{ $\gamma \neq null $} {{$\beta = conflictAnalysis(\cal{F,I,\gamma)}$} \;
        $btl = computeBackjumpLevel(\beta,I)$ \;
        \If{ $btl == 0 $} {\Return UNSAT \;} 
        $\delta = \delta \cup \{\beta\}$ \;
         \If{ $restart()$} {$btl = 0$\;}
        $backjump(btl)$ \;
        $dl = btl$ \;
        }
        \Else{ 
        \If{ $\rho \models \cal{F}$} 			                {$extractPatternFromModel(\rho)$ \;
      $addBlockedClause(\rho)$ \;
      $backjumpUntil(0)$  \;
       \textbf{goto} Prop \;}
       \If{$(timeToReduce())$} {$reduceDB(\delta)$ \;
       $l = selectDecisionVariable(\cal{F})$ \;
       $dl = dl + 1$ \;
      $\rho = \rho$ $\cup$ $\{selectPhase(l)\}$ \;}
        }
      }
 %   }
    %\Return $\Delta$ \;
  }
\end{algorithm}

Algorithm \ref{algoExtendCDCL} depicts the general scheme of CDCL SAT
solver extended for model enumeration. A SAT solver is a tree-based backtrack search procedure; at each node of the search tree, the assigned literals (decision literal and the propagated ones) are labeled with the same decision level starting from 1 and increased at each decision (or branching).

Typically, this solver performs a tree-based backtrack search procedure. Each branch of the binary search tree can be seen as a sequence of decision and unit propagated literals. At each node, a decision variable is chosen (ligne 23), and assigned to the \textit{true} or \textit{false} polarity (\textit{selectPhase(l)} - line 25).
Then unit propagation is performed in line 6. All these literals
(decision and propagated ones) assigned at a given node are labelled with the same level \textit{dl}. If all literals are assigned without contadiction, then $\rho$ is a model of $\cal{F}$ and the formula is answered to be satisfiable (line 16). As our boolean formula represents the encoding of the closed frequent gradual itemset mining problem, each time a model is found, an gradual itemset is extracted from $\rho$ (line 17).
For model enumeration, the search continue by adding a blocked clause to avoid enumerating again the same models (line 18).
Search restart at level 0, to search for the next models (lines 19-20). The other case, is reached when unit propagation (lines 8-14) leads to a conflict ($\gamma$ is the conflict clause), a new asserting clause $\beta$ is derived by conflict analysis (line 8), mostly following the First-UIP
scheme ('Unique Implication Point' \cite{Zhang01})
A backtrack level (\textit{btl}) is derived from the asserting clause (line 9). If \textit{btl} is null, then the formula is answered unsatisfiable (line 10), otherwise $\beta$ is added to the learnt clauses database (line 11) and the algorithm backjump to the level \textit{btl} (line 13).
Regularly, the CDCL solver performs restarts, by backtracking to level 0 (line 12) using one of the various restart strategies (\cite{Huang-07}). Such restarts define the frequency used by the solver to restart the search. Finally, another component concern the learnt clauses management policy. To maintain a learnt clauses database of reasonable size, a reduction is performed (line 22) using one the various strategies proposed in the literature \cite{AudemardS09,MiniSat03,LonlacN17,JabbourLSS14}.

%\section{A SAT-Based Approach for Mining Frequent Gradual Patterns}
%\section{SAT encoding for the frequent gradual pattern mining problem}
\section{SAT-based encoding for the problem of discovering frequent gradual patterns}
\label{encodingSAT}
In this section, we show how the problem of mining all the frequent gradual itemset in a numerical dataset with respect to a minimum support threshold \textit{minSupp}  describe in section \ref{sec:gradual} can be encoded as a boolean formula in $CNF$. Our SAT encoding is inspired on the encodings proposed in \cite{JabbourSS13}.

In oder to formally describe our encoding, we consider a numerical dataset $\Delta = \cal{T} \times \cal{A}$ where $\cal{A}$ $= \{a_1, . . ., a_m\}$ is a set of attributes, $\cal{T}$ $= \{t_1, . . ., t_m\}$ a set of transactions and a minimum support threshold \textit{minSupp}. In the follow, we denote by a parameter $k$ the minimum support threshold. For reasons of clarity, the comparison operator "$\leq$" (respectively "$\geq$") will be denoted "$+$" (respectively "$-$"). We denote by $\cal{A^*}$ the set of attributes variations: 
$\cal{A^*}$ = $\{ a_{1}^+, a_{1}^-, . . ., a_{m}^+, a_{m}^- \}$.
The SAT encoding of frequent gradual itemset mining that we propose is based on the use of propositional variables representing the items and the transaction identifiers in $\Delta$ 

Let ${\cal L} = \langle t_{i_1}, t_{i_2} \ldots t_{i_k}\rangle$ the sequence ordering of the $k$ first transactions as should be appear in the longest sequence of transactions required for a frequent gradual itemset. We denote by $y_{ij}$ the fact that the transaction $t_i$ is set on the $j$th position of $\cal L$.

First of all, we associate with each gradual attribute $a$ two boolean variables respectively  $x_{a^{+}}$  and $x_{a^{-}}$.

The first constraint (\ref{eq:eq1}) allows to not consider  gradual itemset  involving both $a^{+}$ and $a^{-}$ of each attribute $a$.
\begin{equation}  \label{eq:eq1}
	\bigwedge_{a \in a_1..a_n} (\neg x_{a^{+}} \vee \neg x_{a^{-}}) 
\end{equation}

\textit{This first constraint solves the problem encountered with the specialized algorithm  of frequent gradual itemsets mining \texttt{GLCM} \cite{DoTLNTA15} which often returns the gradual itemsets containing both the gradual items and their corresponding complementary gradual items.}

The second constraint (\ref{eq:eq2}) allows us to place  uniquely one transaction $t_i$ in the  $j$th position of a gradual itemset $s$. To this end, a new boolean variable $y_{ij}$ is added to indicate that the transaction $t_i$ is putted in the position $j$.
\begin{equation} \label{eq:eq2}
	\bigwedge_{1 \le j \le k} (\sum_{i=1}^{n} y_{ij} = 1)
\end{equation}

Constraint (\ref{eq:eq3}) is introduced to not allow a transaction to be placed in more than one position in $s$.

\begin{equation} \label{eq:eq3}
	\bigwedge_{1 \le i \le n}  (\sum_{j=1}^{k} y_{ij} \le 1) 
\end{equation}

Constraint (\ref{eq:eq4})  aims to express given a gradual item $a$, the set of transactions that can be set in position $j$ (respectively) those cannot be set constraint (\ref{eq:eq5})).

%\begin{equation} \label{eq:eq4}
%	\bigwedge_{1 \le j \le k}  \bigwedge_{a^{\diamond} \in a_1^+a_1^-..a_n^+a_n^-}  (x_{a^{\diamond}}   \rightarrow  \bigvee_{t_k \in S_j(a^{\diamond})} y_{kj})
%\end{equation}
%
%\begin{equation} \label{eq:eq5a}
%	\bigwedge_{1 \le j \le k}  \bigwedge_{a^{\diamond} \in a_1^+a_1^-..a_n^+a_n^-}  (x_{a^{\diamond}}   \rightarrow  \bigwedge_{t_k \not \in S_j(a^{\diamond})} \neg y_{kj})
%\end{equation}
%
%where $S_j(a^{\diamond}) = \{t_i  \in T~|~ \exists S \subseteq (T\setminus t_k) ~where~ |S| = k-j,  \forall t_k \in  S, ~  t_i(a)\preceq_{\diamond} t_k(a)\}$. $S_j(a^{\diamond})$ is the set of transactions  $t_k$ having at least $k-j$ less than $t_k$ with respect to $\preceq_{\diamond}$.

%Finally, in order to eliminate symmetrical gradual itemsets, we add the following constraint:
%
%\begin{equation}
%\label{eq:symm}
%\bigwedge_{a_i \in a_1.. a_n}  (\neg x_{a_i^+}  \vee \bigvee_{1 \le j <  i} {\neg x_{a_i^-}})
% \end{equation}
%
%In fact, each $\sigma = (a_1^+, a_1^-)\ldots (a_n^+, a_n^-)$ is a symmetry of the our encoding. Consequently, one can break such symmetry by adding the Symmetry Breaking Predicates \cite{?}.

%\newpage

%\textcolor{red}{
\begin{equation} \label{eq:eq4}
	\bigwedge_{a^{\diamond} \in \cal A^{*}} \bigwedge_{1 \le i \le n} \bigwedge_{1 \le j \le k}  (x_{a^{\diamond}} \wedge y_{ij}  \rightarrow  \bigvee_{t_k(a) ~\diamond~ t_i(a)} y_{k(j+1)})
\end{equation}
Note that such constraint can be expressed differently by considering only those that are not allowed as stated in Constraint (\ref{eq:eq5}). In contrast to (\ref{eq:eq4}), constraint (\ref{eq:eq5}) allows to add only ternary clauses. However, their number is higher that those of (\ref{eq:eq4}).
\begin{equation} \label{eq:eq5}
	\bigwedge_{a^{\diamond} \in \cal A^{*}} \bigwedge_{1 \le i \le n} \bigwedge_{1 \le j \le k}  (x_{a^{\diamond}} \wedge y_{ij}  \rightarrow  \bigwedge_{t_k(a) ~\overline{\diamond}~ t_i(a)} \neg y_{k(j+1)})
\end{equation}
\\
Finally, in order to eliminate symmetrical gradual itemsets, we add the following constraint:
\\
\begin{equation}
\label{eq:symm}
\bigwedge_{a_i \in a_1.. a_n}  (\neg x_{a_i^+}  \vee \bigvee_{1 \le j <  i} {\neg x_{a_i^-}})
 \end{equation}
\\
In fact, each $\sigma = (a_1^+, a_1^-)\ldots (a_n^+, a_n^-)$ is a symmetry of the our encoding. Consequently, one can break such symmetry by adding the Symmetry Breaking Predicates \cite{crawford96c}.
\\
Note that the equation (\ref{eq:eq4}) or (\ref{eq:eq5}) can be simplified to ($\neg x_{a^{\diamond}} \vee \neg y_{ij}$) when it is not possible to attribute transactions to positions  between $j+1$ to $k$ with transactions allowing to maintain the relation ($\diamond$) between positions $j+1$ and $k$. This is  the case if $(j-1) < |\{l~|~t_i(a) \diamond t_l(a)\}|$ or $(k-j) < |\{l~|~t_l(a) ~\diamond~ t_i(a)\}|$. Note that the computation of $|\{l~|~t_l(a) ~\diamond~ t_i(a)\}|$ can be done by double traversal of the transaction database ${\Delta}$.
Such processing allows to reduce the number of added clauses if constraint  (\ref{eq:eq5}) is used while it allows to reduce the size of added clauses if  (\ref{eq:eq4}) is used.
Note that $(\sum_{i=1}^{n} y_{ij} = 1)$ (respectively  $(\sum_{j=1}^{k} y_{ij} \le 1)$) represent linear equality (respectively inequality) commonly called exact-One (respectively atMostOne Constraint). Such constraint can be encoding in respectively $O(n)$ (respectively  $O(k)$) clauses using $O(n)$ (respectively  $O(k)$)  additional variables as indicated in constraint (\ref{eq:atMostOne}) \cite{WARNERS199863,SilvaL07}. In fact, $\sum_{i=1}^{n} x_i \le 1$ can be encoded as follows using auxiliary variables $\{p_1, \ldots, p_{n-1}\}$.
\begin{equation}
\label{eq:atMostOne}
 (\neg x_1 \vee p_1) \wedge ( \neg x_n \vee \neg p_{n-1}) \wedge \bigwedge_{1 < i < n} (\neg x_i \vee p_i)\wedge (\neg p_{i-1} \vee p_i) \wedge (\neg x_i \vee \neg p_{i-1}) 
 \end{equation}
%}

\subsection{Adding multiple constraints}

The constraint (\ref{eq:symm}) allows to avoid computing all gradual patterns and their corresponding symmetric gradual pattern. However, this constraint will add a certain number of variables and clauses to the final boolean formula.
We propose another direction to take into account this symmetrical without add the constraint (\ref{eq:symm}) but by
adding two blocking clauses in the NCF formula each time a model is found. One clause to avoid finding the same model and another to avoid finding a model corresponding to the symmetric pattern.
%the variable polarity selection heuristic \cite{JeroslowW90}.
%In fact, during the resolution process of the CNF formula encoding the frequent gradual pattern mining problem, we always affect to the first branching variable a positive polarity.
%In this way, all the models of the obtained CNF formula will have their first variable with positive polarity.

%\textcolor{red}{
Several other constraints over the pattern itself can be captured by the variable selection heuristic.
In many application fields, interesting gradual patterns can be
distinguished from irrelevant ones by specifying semantic
constraints on the gradual pattern itself. For example, the authors of \cite{LonlacR17} designed an algorithm to mine temporal gradual patterns which are gradual patterns whose the longest sequence of transactions respect the temporal order.
These kinds of gradual patterns are particularly interesting in the paleoecological domain where the experts search from their paleoecological numerical data the patterns which capture the simultaneously frequent co-evolutions between attributes.
As the transactions are encoded in our CNF formula as boolean variables, the temporal constraint can be captured by selecting in the temporal order the propositional variables $y_{ij}$ representing the transaction identifiers of the numerical dataset.
%}. 

\subsection{Solving the formula encoding gradual pattern mining problem}
We solve our SAT boolean formula using $MiniSAT 2.2$ CDCL SAT solver \cite{MiniSat03}.
Each model of our SAT formula (i.e., each solution) is a frequent gradual pattern of the input database with respect to a minimum support threshold. As a result, outputting all the frequent gradual pattern can be done by enumerating all the models that satisfy the CNF formula encoding the frequent gradual pattern mining problem.

The main procedure of our approach is given in algorithm \ref{algoExtendCDCL_forGradual}. This procedure compute and output all frequent gradual patterns with respect to the minimum support threshold \textit{minSupp}.

The procedure  \textit{findAllModel} corresponds to the algorithm \ref{algoExtendCDCL} modified by  adding to the CNF formula two blocking clauses instead of one blocking clause at each time that a model is found  during the resolution process. One blocking clause to avoid finding the same model and another to avoid finding a model corresponding to the symmetric pattern of the extracted gradual pattern.
More precisely, let $s = (a_{i1}^{i1}, a_{i2}^{i2}, \cdots, a_{ik}^{ik})$, a frequent gradual pattern extracted from the current model, we add to the original formula the blocking clauses: $c_1 = (\neg x_{a_{i1}^{i1}} \vee \neg x_{a_{i2}^{i2}} \vee  \cdots \vee \neg x_{a_{ik}^{ik}})$ and $c_2 = (\neg x_{a_{j1}^{j1}} \vee \neg x_{a_{j2}^{j2}} \vee  \cdots \vee \neg x_{a_{jk}^{jk}})$, with $ a_{j1}^{j1} = c(a_{i1}^{i1}), a_{j2}^{j2} = c(a_{i2}^{i2}), \dots, a_{jk}^{jk} = c(a_{ik}^{ik})$. The clause $c_2$ allows to discard from the set of patterns the complementary gradual pattern of $s$.

\begin{algorithm}[H]
  \caption{SAT Based Gradual Patterns Enumeration}
  \label{algoExtendCDCL_forGradual}
  %\SetVline
  \SetInd{0.3em}{0.7em}
  \KwIn{ a numerical database $\cal{DS}$, a minimum support \textit{minSupp}}
  \KwOut{Set of all frequent gradual patterns}
  
  \SetKwBlock{Begin}{Debut}{Fin}  
  {
   % \If{\TIMEREDUCE{}}{
      $\cal{F}$ $\leftarrow$ $SATEncoding(DS, \textit{minSupp})$ \;
      $findAllModel$($\cal{F}$) \;
      
%       \While{$\cal{F}$ is solvable}
%       {
%       	model $\leftarrow$  $nextModel(\cal{F})$  \;
%       }
 %   }
    %\Return $\Delta$ \;
  }
\end{algorithm}

\section{Experiments}
\label{exps}
In this section, we carried out an experimental evaluation of
the performance of our proposed approach. 
we ran experiments on the paleoecological datasets.
The paleoecological dataset are constituted of a set of numerical attributes whose the values correspond to the quantity of each paleoecological indicator contained in a sediment record taken, by coring operations, in a lake ecosystem.
The sedimentary sequence obtained is then dated, sampled, and for each sample, at a given depth, a date is calculated.  The abundance of each indicator is then recorded for each sample. The objects in this database correspond to the different dates obtained on the considered sedimentary record, and the columns to the different paleoecological recorded.
We consider the paleoecological dataset constituted of the indicators of paleoecological anthropization (pollen grains). 
It contains $111$ objects corresponding to different dates identified on the considered Lacustrine recording, and $117$ attributes corresponding to different indicators of paleoecological anthropization (pollen grains).
All the experiments were done on Intel Xeon quad-core
machines with 32GB of RAM running at 2.66 Ghz.
First, we present in the table \ref{tab:sizeCNF} the size of the CNF formula (number of variables and clauses) encoding the frequent gradual patterns with respect to a minimum support.

\begin{table}[!h]
\begin{center}
%\begin{Large}
\renewcommand{\arraystretch}{1.4}
\caption{CHARACTERISTICS OF THE INSTANCES \&  ENCODING TIME}
\label{tab:sizeCNF}  
\begin{tabular}{cccc}
  \hline
\textbf{~~~~\#minSupp}~~~~ & \textbf{~~~~\#vars~~~~} & \textbf{~~~~\#clauses~~~~} & \textbf{~~~~\#encodingTime~~~~} \\
\hline 
$5\%$ & 2 115 & 133 521 & 0.22s \\
\hline 
$10\%$ & 3 775 & 266 706 & 0.43s \\
\hline
$20\%$ & 7 427 &  559 713 & 0.86s \\
  \hline
$30\%$ & 11 079 &  852 720 & 1.31s \\
\hline
$40\%$ & 14 731 & 1 145 727 &  1.74s \\
\hline
$50\%$ & 18 383 & 1 438 734 & 2.25s \\
\hline
$60\%$ & 22 035 & 1 731 741  & 2.69s \\
\hline
$70\%$ & 25 687 & 2 024 748 & 3.12s \\
\hline
$80\%$ & 29 339 & 2 317 755  & 3.54s \\
\hline
$90\%$ & 32 991 & 2 610 762  & 4.03s \\
\hline
\hline
\end{tabular}
%\end{Large}
\end{center}
\end{table}

In this table, we mention the formula encoding the whole problem in terms of number of variables $(\#vars)$ and clauses $(\#clauses)$ with respect to a minimum support threshold $(\#minSupp)$ given by the first column. The last column gives in seconds the cpu time need for encoding.

\section{Conclusion}
In this paper, we proposed SAT encoding to address the problem of mining frequent gradual patterns. This declarative approach offers an additional possibility to benefit from the recent progress in satisfiability testing.
Several satisfiability based approach have been proposed for the classical patterns mining problem such that mining frequent itemsets in transactional
data, mining frequent sequence in a data-sequence. However no satisfiability based approach has yet been proposed for the frequent gradual pattern mining problem. The problem of mining frequent gradual patterns differs from the classical cases related to simple itemsets. In fact, in this last case, for each line of the database, it is possible to say whether it supports the given itemset or not.
In the gradual case, the entire database is needed for each count.
This makes the problem of mining frequent gradual patterns more complex.

\bibliographystyle{plain}
\bibliography{biblio,satBib}
\end{document}

%% file: symbols.tex
\def\infty{\text{BUG}}

\renewcommand{\leq}{\mathrel{\leqslant}}
\renewcommand{\geq}{\mathrel{\geqslant}}

%% file: main.bbl
\begin{thebibliography}{10}

\bibitem{AgrawalS94}
Rakesh Agrawal and Ramakrishnan Srikant.
\newblock Fast algorithms for mining association rules in large databases.
\newblock In {\em VLDB'94, Proceedings of 20th International Conference on Very
  Large Data Bases, September 12-15, 1994, Santiago de Chile, Chile}, pages
  487--499, 1994.

\bibitem{AudemardS09}
G.~Audemard and L.~Simon.
\newblock Predicting learnt clauses quality in modern sat solvers.
\newblock In {\em Proceedings of the 21st International Joint Conference on
  Artificial Intelligence, IJCAI'09}, pages 399--404, 2009.

\bibitem{AyouniLYP10}
Sarra Ayouni, Anne Laurent, Sadok~Ben Yahia, and Pascal Poncelet.
\newblock Mining closed gradual patterns.
\newblock In {\em Artificial Intelligence and Soft Computing, 10th
  International Conference, {ICAISC} 2010, Zakopane, Poland, June 13-17, 2010,
  Part {I}}, pages 267--274, 2010.

\bibitem{BerzalCSVS07}
Fernando Berzal, Juan~C. Cubero, Daniel S{\'{a}}nchez, Mar{\'{\i}}a Amparo~Vila
  Miranda, and Jos{\'{e}}{-}Mar{\'{\i}}a Serrano.
\newblock An alternative approach to discover gradual dependencies.
\newblock {\em International Journal of Uncertainty, Fuzziness and
  Knowledge-Based Systems}, 15(5):559--570, 2007.

\bibitem{BiereMH2009}
Armin Biere, Marijn Heule, Hans van Maaren, and Toby Walsh, editors.
\newblock {\em Handbook of Satisfiability}, volume 185 of {\em Frontiers in
  Artificial Intelligence and Applications}. {IOS} Press, 2009.

\bibitem{CoqueryJSS12}
Emmanuel Coquery, Sa\"{\i}d Jabbour, Lakhdar Sa\"{\i}s, and Yakoub Salhi.
\newblock A sat-based approach for discovering frequent, closed and maximal
  patterns in a sequence.
\newblock In {\em Proceedings of the 20th European Conference on Artificial
  Intelligence (ECAI'12)}, pages 258--263, 2012.

\bibitem{crawford96c}
James Crawford, Matthew~L. Ginsberg, Eugene Luck, and Amitabha Roy.
\newblock Symmetry-breaking predicates for search problems.
\newblock In {\em Principles of Knowledge Representation and Reasoning
  (KR'96)}, pages 148--159. 1996.

\bibitem{Di-JorioLT08}
Lisa Di{-}Jorio, Anne Laurent, and Maguelonne Teisseire.
\newblock Fast extraction of gradual association rules: a heuristic based
  method.
\newblock In {\em {CSTST} 2008: Proceedings of the 5th International Conference
  on Soft Computing as Transdisciplinary Science and Technology,
  Cergy-Pontoise, France, October 28-31, 2008}, pages 205--210, 2008.

\bibitem{Di-JorioLT09}
Lisa Di{-}Jorio, Anne Laurent, and Maguelonne Teisseire.
\newblock Mining frequent gradual itemsets from large databases.
\newblock In {\em Advances in Intelligent Data Analysis VIII, 8th International
  Symposium on Intelligent Data Analysis, {IDA} 2009, Lyon, France, August 31 -
  September 2, 2009. Proceedings}, pages 297--308, 2009.

\bibitem{DoTLNTA15}
Trong Dinh~Thac Do, Alexandre Termier, Anne Laurent, Benjamin
  N{\'{e}}grevergne, Behrooz~Omidvar Tehrani, and Sihem Amer{-}Yahia.
\newblock {PGLCM:} efficient parallel mining of closed frequent gradual
  itemsets.
\newblock {\em Knowl. Inf. Syst.}, 43(3):497--527, 2015.

\bibitem{DuboisP92a}
Didier Dubois and Henri Prade.
\newblock Gradual inference rules in approximate reasoning.
\newblock {\em Inf. Sci.}, 61(1-2):103--122, 1992.

\bibitem{DuboisPU03}
Didier Dubois, Henri Prade, and Laurent Ughetto.
\newblock A new perspective on reasoning with fuzzy rules.
\newblock {\em Int. J. Intell. Syst.}, 18(5):541--567, 2003.

\bibitem{MiniSat03}
Niklas E{\'e}n and Niklas S{\"o}rensson.
\newblock An extensible sat-solver.
\newblock In {\em Proceedings of the Sixth International Conference on Theory
  and Applications of Satisfiability Testing (SAT'03)}, pages 502--518, 2003.

\bibitem{EenS03}
N.~En and N.~S{\"o}rensson.
\newblock An extensible \protect{SAT}-solver.
\newblock pages 502--518, 2003.

\bibitem{Huang-07}
Jinbo Huang.
\newblock The effect of restarts on the efficiency of clause learning.
\newblock pages 2318--2323.

\bibitem{Hullermeier02}
Eyke H{\"{u}}llermeier.
\newblock Association rules for expressing gradual dependencies.
\newblock In {\em Principles of Data Mining and Knowledge Discovery, 6th
  European Conference, {PKDD} 2002, Helsinki, Finland, August 19-23, 2002,
  Proceedings}, pages 200--211, 2002.

\bibitem{JabbourLSS14}
Sa{\"{\i}}d Jabbour, Jerry Lonlac, Lakhdar Sais, and Yakoub Salhi.
\newblock Revisiting the learned clauses database reduction strategies.
\newblock {\em CoRR}, abs/1402.1956, 2014.

\bibitem{JabbourSS13}
Sa{\"{\i}}d Jabbour, Lakhdar Sais, and Yakoub Salhi.
\newblock The top-k frequent closed itemset mining using top-k {SAT} problem.
\newblock In {\em Machine Learning and Knowledge Discovery in Databases -
  European Conference, {ECML} {PKDD} 2013, Prague, Czech Republic, September
  23-27}, pages 403--418, 2013.

\bibitem{KendallB39}
Michel Kendall and Babington Smith.
\newblock The problem of m rankings.
\newblock In {\em The annals of mathematical statistics - Volume 10}, pages
  275--287, 1939.

\bibitem{LaurentLR09}
Anne Laurent, Marie{-}Jeanne Lesot, and Maria Rifqi.
\newblock {GRAANK:} exploiting rank correlations for extracting gradual
  itemsets.
\newblock In {\em Flexible Query Answering Systems, 8th International
  Conference, {FQAS} 2009, Roskilde, Denmark, October 26-28, 2009.
  Proceedings}, pages 382--393, 2009.

\bibitem{LaurentNST10a}
Anne Laurent, Benjamin N{\'{e}}grevergne, Nicolas Sicard, and Alexandre
  Termier.
\newblock Efficient parallel mining of gradual patterns on multicore
  processors.
\newblock In {\em Advances in Knowledge Discovery and Management - Volume 2
  [Best of {EGC} 2010, Hammamet, Tunisie]}, pages 137--151, 2010.

\bibitem{LonlacN17}
Jerry Lonlac and Engelbert {Mephu Nguifo}.
\newblock Towards learned clauses database reduction strategies based on
  dominance relationship.
\newblock {\em CoRR}, abs/1705.10898, 2017.

\bibitem{LonlacR17}
Jerry Lonlac, Yannick Miras, Aude Beauger, Marie Pailloux, Jean-Luc Peiry, and
  Engelbert~Mephu Nguifo.
\newblock Une approche d'extraction de motifs graduels (fermés) fréquents
  sous contrainte de la temporalité.
\newblock {\em Revue des Nouvelles Technologies de l'Information}, Extraction
  et Gestion des Connaissances, RNTI-E-33:213--224, 2017.

\bibitem{Masseglia08gradualtrends}
Florent Masseglia, Anne Laurent, and Maguelonne Teisseire.
\newblock Gradual trends in fuzzy sequential patterns.
\newblock In {\em In IPMU}, pages 456--463, 2008.

\bibitem{Moskewicz01}
M.~W. Moskewicz, C.~F. Madigan, Y.~Zhao, L.~Zhang, and S.~Malik.
\newblock Chaff: Engineering an efficient \protect{SAT} solver.
\newblock In {\em Proceedings of the 38th Design Automation Conference
  ({DAC}'01)}, pages 530--535, 2001.

\bibitem{PasquierBTL99}
Nicolas Pasquier, Yves Bastide, Rafik Taouil, and Lotfi Lakhal.
\newblock Efficient mining of association rules using closed itemset lattices.
\newblock {\em Inf. Syst.}, 24(1):25--46, 1999.

\bibitem{SilvaL07}
Jo{\~a}o P.~Marques Silva and In{\^e}s Lynce.
\newblock Towards robust cnf encodings of cardinality constraints.
\newblock In {\em CP}, pages 483--497, 2007.

\bibitem{SrikantA96}
Ramakrishnan Srikant and Rakesh Agrawal.
\newblock Mining quantitative association rules in large relational tables.
\newblock In {\em Proceedings of the 1996 {ACM} {SIGMOD} International
  Conference on Management of Data, Montreal, Quebec, Canada, June 4-6, 1996.},
  pages 1--12, 1996.

\bibitem{Tseitin68}
G.S. Tseitin.
\newblock On the complexity of derivations in the propositional calculus.
\newblock In H.A.O. Slesenko, editor, {\em Structures in Constructives
  Mathematics and Mathematical Logic, Part II}, pages 115--125, 1968.

\bibitem{WARNERS199863}
Joost~P. Warners.
\newblock A linear-time transformation of linear inequalities into conjunctive
  normal form.
\newblock {\em Information Processing Letters}, 68(2):63 -- 69, 1998.

\bibitem{Zhang01}
L.~Zhang, C.~F. Madigan, M.~W. Moskewicz, and S.~Malik.
\newblock Efficient conflict driven learning in \protect{Boolean}
  satisfiability solver.
\newblock In {\em IEEE/ACM CAD'2001}, pages 279--285, 2001.

\end{thebibliography}
